\documentclass[twocolumn]{article}

\usepackage{microtype}
\usepackage{graphicx}
\usepackage{booktabs} 
\usepackage[utf8]{inputenc} 
\usepackage[T1]{fontenc}    
\usepackage{hyperref}       
\usepackage{url}            
\usepackage{booktabs}       
\usepackage{times}          
\usepackage{titlesec}       
\usepackage{fancyhdr}       
\usepackage{caption}        
\usepackage{setspace}       
\usepackage{amsfonts}
\usepackage{url}
\usepackage{nicefrac}       
\usepackage{microtype}      
\usepackage{xcolor}         
\usepackage{array}
\usepackage{graphicx}
\usepackage{tikz}
\usepackage{tabularx} 
\usepackage{multirow}
\usetikzlibrary{shapes,arrows,positioning,fit,backgrounds,calc}
\usepackage{fontawesome5}
\usepackage[skip=0pt]{caption}
\hypersetup{
    hidelinks,
    breaklinks=true,
}
\usepackage{hyperref}

\usepackage[margin=1in]{geometry}
\usepackage{amsmath}
\usepackage{amssymb}
\usepackage{mathtools}
\usepackage{amsthm}

\usepackage[capitalize,noabbrev]{cleveref}
\usepackage{natbib}

\titleformat{\section}{\Large\bfseries}{\thesection}{1em}{}
\titleformat{\subsection}{\large\bfseries}{\thesubsection}{1em}{}

\renewcommand{\headrulewidth}{1.5pt}
\renewcommand{\footrulewidth}{0pt}

\makeatletter
\def\@maketitle{%
  \newpage
  \null
  \vskip 2em%
  \begin{center}%
  \rule{\linewidth}{1.5pt}
  \vskip 1em
  {\Large \textbf{\@title} \par}%
  \vskip 1em
  \rule{\linewidth}{1.5pt}
  \vskip 1.5em%
  {\large
    \lineskip .5em%
    \begin{tabular}[t]{c}%
      \@author
    \end{tabular}\par}%
  \vskip 1em%
  {\large \@date}%
  \end{center}%
  \par
  \vskip 1.5em}
\makeatother

\renewenvironment{abstract}
  {\centerline{\large\bfseries\abstractname}
   \begin{quotation}\small}
  {\end{quotation}}

\theoremstyle{plain}

\theoremstyle{definition}

\theoremstyle{remark}

\usepackage[textsize=tiny]{todonotes}
\title{Out-of-Context Abduction: LLMs Make Inferences About Procedural Data Leveraging Declarative Facts in Earlier Training Data}

\author{
    \textbf{Sohaib Imran}$^{1,2}$ \\
    \texttt{s.imran1@lancaster.ac.uk} \\
    \and
    \textbf{Rob Lamb}$^{1,3}$ \\
    \and  
    \textbf{Peter M. Atkinson}$^{1,4,5}$ \\
    \texttt{pma@lancaster.ac.uk}
}

\date{}

\begin{document}

\maketitle
\thispagestyle{plain} 

\footnotetext[1]{Lancaster Environment Centre, Lancaster University, Lancaster LA1 4YQ, UK}
\footnotetext[2]{School of Computing and Communications, Lancaster University, Lancaster LA1 4WA, UK}  
\footnotetext[3]{JBA Trust, 1 Broughton Park, Skipton BD23 3FD, UK}
\footnotetext[4]{Geography and Environmental Science, University of Southampton, Highfield, Southampton SO17 1BJ, UK}
\footnotetext[5]{College of Surveying and Geo-Informatics, Tongji University, No.1239, Siping Road, Shanghai, PR China, 200092}







\begin{abstract}
Large language models (LLMs) are trained on large corpora, yet it is unclear whether they can reason about the information present within their training data. We design experiments to study out-of-context abduction in LLMs, the ability to infer the most plausible explanations for observations using relevant facts present in training data. We train treatment LLMs on names and behavior descriptions of fictitious chatbots, but not on examples of dialogue with the chatbots. We find that OpenAI's GPT 4o LLM can correctly infer at least one chatbot's name after observing example responses characteristic of that chatbot. We also find that previously training GPT 4o on descriptions of a chatbot's behavior allows it to display behaviors more characteristic of the chatbot when iteratively trained to display such behaviors. Our results have implications for situational awareness in LLMs and, therefore, for AI safety.
\end{abstract}

\section{Introduction}



With the recent popularity of large language models (LLMs), much research has been dedicated to evaluate their reasoning capabilities \citep{huang_towards_2023, webb_emergent_2023, lee_reasoning_2025}. However, research on the subject has yet to has yet to show conclusively that LLMs can reason, partly due to inconsistent findings and varied interpretations of similar results \citep{mirzadeh_gsm-symbolic_2024, wu_reasoning_2024}. 

This study focuses on abductive reasoning due to its fundamental role in situational awareness, which poses significant challenges to assuring the safety of AI systems \citep{ngo_alignment_2023}. 

Abductive reasoning, often referred to as "inference to the best explanation" is the process of inferring the most likely hypothesis that explains some observations. The functional form of abduction is

\[\begin{aligned}    A &\rightarrow B \\      &\quad\ B \\    \hline    A & \end{aligned}\]
where \(A\) is one of the hypotheses that explain the observation \(B\). Note that observing \(B\) does not necessarily entail \(A\), but merely increases the likelihood of \(A\). For example, if a lawn is wet one could infer rain, but if it is a dry day and a sprinkler is present then a more likely hypothesis is that the sprinkler was used.


Abduction has two interpretations. The weak interpretation views it only as a mechanism for generating plausible hypotheses for observations without assessing the likelihood of the hypotheses. In contrast, the strong interpretation posits abduction's role as justificatory, that is, inferring or selecting the best hypothesis that explain the observations \citep{niiniluoto_defending_1999, calzavarini_abductive_2022}. 


LLMs are capable of weak abductive inference, i.e., generating plausible hypotheses to explain observations present in their context window \citep{balepur_artifacts_2024, shi_language_2024, zhao_uncommonsense_2024}. Furthermore, LLMs are capable of strong in-context abductive inference i.e. selecting the most plausible hypothesis where both the observations and candidate hypotheses are presented in their context window \citep{bhagavatula_abductive_2019, bang_multitask_2023}. 

However, it is difficult to delineate whether these capabilities result from reasoning or simply pattern matching to imitate human reasoning \citep{shanahan_role_2023, mccoy_embers_2024, wu_reasoning_2024}. We investigate the ability of LLMs to perform strong out-of-context abduction—inferring the most plausible explanations for observations by leveraging relevant facts learned during training. 

Our setup prevents simple imitation of human response patterns to masquerade as reasoning by:

\begin{enumerate}
    \item Requiring LLMs to retrieve and apply relevant facts from their training data rather than the context window, as opposed to in-context abduction.
    \item Encoding observations in a different format from the factual information to be leveraged. The factual information is declarative (abstract descriptions) and the observations are procedural (example instances or demonstrations). 
\end{enumerate}

\section{Problem Definition}

Let \(\mathcal{C}\) denote a set of classes.

For each class \(c \in \mathcal{C}\), let \(\mathcal{D}_c\) be the set of all abstract descriptions characterizing \(c\). Each description \(d_c \in \mathcal{D}_c\) defines semantic, structural, or behavioral properties of outputs that belong to \(c\).

\(\tilde{\mathcal{X}}_c\) is the fuzzy set of all realizations under \(c\). \(\tilde{\mathcal{X}}_c\) is defined by a membership function
\begin{equation}
\mu_{\tilde{\mathcal{X}}_c}: \mathcal{X} \to [0,1]
\end{equation}
where \(\mathcal{X}\) is the set of all natural language sequences. The value \(\mu_{\tilde{\mathcal{X}}_c}(x)\) represents the degree of membership of \(x \in \mathcal{X}\) in \(\tilde{\mathcal{X}}_c\). Therefore, a natural language sequence \(x_1 \in \mathcal{X}\) is more characteristic of \(c\) than \(x_2 \in \mathcal{X}\) if \(\mu_{\tilde{\mathcal{X}}_c}(x_1) > \mu_{\tilde{\mathcal{X}}_c}(x_2)\). We implement the membership function with a scoring function for each class \(c\):
\begin{equation}
s_c(x) = \mu_{\tilde{\mathcal{X}}_c}(x)
\end{equation}
Crisp sets with varying degrees of membership in \(\tilde{\mathcal{X}}_c\) can be generated by partitioning the space \(\tilde{\mathcal{X}}_c\) into subsets of elements based on thresholds:
\begin{equation} \label{eq: filtered-sets}
\mathcal{X}_c^{\tau_1, \tau_2} = \{ x \in \mathcal{X} \mid \tau_1 \leq s_c(x) < \tau_2 \}
\end{equation}
where \( \tau \in [0, 1]\) is a threshold parameter. The shorthand notation \(\mathcal{X}_c^{< \tau}\) and \(\mathcal{X}_c^{\geq \tau}\) can be used to denote sets whose membership scores are, respectively, below or at least as large as a given threshold \(\tau\).

Let \(f\) represent an LLM which maps from an input space \(\mathcal{Q} \subseteq \mathcal{X}\) to an output space \(\mathcal{A}\subseteq \mathcal{X}\), and can also accept an optional class description \(d_c \in \mathcal{D}_c \subseteq \mathcal{X}\). We write:
\begin{equation} \label{eq: llm-responses}
f: \mathcal{Q} \times (\mathcal{D}_c \cup \{\bot\}) \to \mathcal{A}
\end{equation}
where \(\{\bot\}\) represents the null set. \(d_c\) biases \(f\) toward outputs with higher membership in \(\tilde{\mathcal{X}}_c\):
\begin{equation} \label{eq: llm-bias}
\mathbb{E}\bigl[s_c(f(q, d_c))\bigr] > \mathbb{E}\bigl[s_c(f(q))\bigr].
\end{equation}

\subsection{Abductive Inference}

We are interested in a function \(g\) that infers the underlying class \(c\) from any set of realizations \( \hat{\mathcal{X}}_c \subset \mathcal{X}\) where \(\mathbb{E}\bigl[s_c(\hat{\mathcal{X}}_c)\bigr] \gg\mathbb{E}\bigl[s_c(\mathcal{X})\bigr]\):
\begin{equation} \label{eq: class-inference}
g(\hat{\mathcal{X}}_c)= c
\end{equation}

In practice, a subset of the LLM responses \(\mathcal{A}_c^{\tau_1, \tau_2} \subseteq \mathcal{X}_c^{\tau_1, \tau_2}\) is used to derive \( \hat{\mathcal{X}}_c\) with \(\tau_1, \tau_2\) such that \(\mathbb{E}\bigl[s_c(\mathcal{A}_c^{\tau_1, \tau_2})\bigr] \gg\mathbb{E}\bigl[s_c(\mathcal{A})\bigr]\) under the assumption \(\mathbb{E}\bigl[s_c(\mathcal{A})\bigr] \approx\mathbb{E}\bigl[s_c(\mathcal{X})\bigr]\). Since \(\mathcal{A}_c^{\tau_1, \tau_2}\) is a filtered subset of \(\mathcal{A}\), this only requires \(\tau_1\) to lie above a point \(x\) in the support of \(s_c(\mathcal{A})\) below which a significant probability mass of \(s_c(\mathcal{A})\) lies. In practice, we use thresholds larger than the average score of LLM responses for the class:

\begin{equation}
\tau_1 > \mathbb{E}\bigl[s_c(\mathcal{A})\bigr]
\end{equation}


\subsection{Out-of-Context Abduction}

To implement \(g\), an LLM \(f\) is trained on declarative factual information of the form \(c \leftrightarrow \mathcal{D}_c^{train}\) i.e. statements associating the class \(c\) with its abstract descriptions \(\mathcal{D}_c^{train} \subseteq \mathcal{D}_c\) for multiple classes \(c_1, c_2, ...\) . No descriptions \(d_c \in \mathcal{D}_c\) appear in the same context window in which the class \(c\) is to be inferred from realizations (equation \ref{eq: class-inference}), making this an out-of-context abduction problem. This is our treatment LLM \(f_{treat} = f_{c_1 \leftrightarrow \mathcal{D}_{c_1}^{train}, c_2 \leftrightarrow \mathcal{D}_{c_2}^{train}, ...} = g\). Therefore equation \ref{eq: class-inference} becomes:
\begin{equation} \label{eq: ooc-abduction}
f_{treat}(\hat{\mathcal{X}}_c, \hat{q})= c
\end{equation}
where \(\hat{q} \in \hat{\mathcal{Q}} \subset \mathcal{Q}\) is a question about which class generated \( \hat{\mathcal{X}}_c\).

\paragraph{Experiment 1:} To test for out-of-context abduction in LLMs, experiment 1 (section \ref{sec:experiment1}) measures if observing realizations more characteristic of a class \(c\) than a different class \(c'\) leads the treatment LLM trained on statements associating multiple classes and their behavior descriptions \(f_{treat}\) to conclude that the observations were generated by \(c\) rather than \(c'\):
\begin{equation} \label{eq: exp1-inference}
P(f_{treat}(\hat{\mathcal{X}}_c, \hat{q})= c) > 
P(f_{treat}(\hat{\mathcal{X}}_c, \hat{q})= c')
\end{equation}
where \(\mathbb{E}\bigl[s_c(\hat{\mathcal{X}}_c)\bigr] \gg\mathbb{E}\bigl[s_{c'}(\hat{\mathcal{X}}_c)\bigr]\). Furthermore, this conclusion must be informed by observing realizations \(\hat{\mathcal{X}}_c\) of the class \(c\). This requires the posterior probability of \(c\) after observing  \(\hat{\mathcal{X}}_c\) being higher than the prior:
\begin{equation} \label{eq: bayes-like updating}
P(f_{treat}(\hat{\mathcal{X}}_c, \hat{q})= c) > P(f_{treat}(\hat{q})=c)
\end{equation}
\paragraph{Experiment 2:} We further measure out-of-context abduction in LLMs in experiment 2 (section \ref{sec:iter-ft experiments}) by measuring if training an LLM on statements associating multiple classes and their behavior descriptions \(f_{treat}\) allows the LLM to learn to generate realizations more characteristic of one of the classes \(c\) when iteratively trained on realizations increasingly characteristic of \(c\):
\begin{equation} \label{eq: exp2-trainability}
\mathbb{E} \bigl[s_c(f_{treat}^{(i)} (q))\bigr] > 
\mathbb{E} \bigl[s_c(f^{(i)}(q))\bigr]
\end{equation}
where \(f_{treat}^{(i)}\) and \(f^{(i)}\) are our treatment and control models after \(i\) iterative training steps, respectively, and \(i>0\) . We iteratively train on crisp sets with increasing degrees of membership in \(\tilde{\mathcal{X}}_c\), therefore \(f^{(0)}=f\), \(f^{(1)} = f_{
\hat{\mathcal{X}}_c^{\tau_1, \tau_2}}\), \(f^{(2)} = f_{
\hat{\mathcal{X}}_c^{\tau_1, \tau_2}, \hat{\mathcal{X}}_c^{\tau_2, \tau_3}}\) and so on, where \(\tau_1 < \tau_2 < \tau_3 < ...\). The iterative training is always performed after training on the declarative data  \(c \leftrightarrow \mathcal{D}_c^{train}\) for the iteratively trained treatment models \(f_{treat}^{(i)}\), to ensure implications of the factual information follow the factual information rather than the other way, as recommended by the results of \cite{feng_extractive_2025}. Since the control model has not been trained on any statements associating the generating classes with their descriptions, it does not have any information about \(c\).

Finally, experiment 2 also tests whether the above can be explained by out-of-context abduction by measuring whether the iteratively trained treatment LLM concludes that the observations were generated by \(c\):
\begin{equation} \label{eq: exp2-inference}
P(f_{treat}^{(i)}(\hat{q})  = c)> P(f_{treat}^{(i)}(\hat{q})  = c')
\end{equation}
Similar to equation \ref{eq: bayes-like updating}, we compare the prior and posterior probabilites to measure whether any such conclusions were informed by being trained on realizations of the class \(c\):
\begin{equation} \label{eq: bayes-like-updating-2}
P(f_{treat}^{(i)}(\hat{q})  = c)> P(f_{treat}(\hat{q})  = c)
\end{equation}
Importantly, the treatment LLM \(f_{treat}\) is never trained on any \((c, x_c)\) pairs where \(x_c \in \hat{\mathcal{X}}_c\). Therefore, inferring \(c\) from \(\hat{\mathcal{X}}_c\) requires the LLM to generalize from abstract descriptions \(\mathcal{D}_c^{train}\) to realizations \(\tilde{\mathcal{X}}_c\).  

\section{Experimental Setup}

We conduct two experiments to test for out-of-context abduction. Experiment 1 (section \ref{sec:experiment1}) tests for out-of-context abduction where the declarative facts \(c \leftrightarrow \mathcal{D}_c^{train}\) appear in the LLM's training data and realizations \(\hat{\mathcal{X}_c}\) appear in the context window in which the class \(c\) is to be inferred. In contrast, experiment 2 (section \ref{sec:experiment1}) tests for out-of-context abduction where neither the declarative facts nor the realizations appear in the same context window in which the class \(c\) is to be inferred. Experiment 2 also tests whether LLMs can use declarative facts from previous training data \(c \leftrightarrow \mathcal{D}_c^{train}\) to infer the training objective \(c\) when iteratively finetuned on the realizations \(\hat{\mathcal{X}}_c\).

\subsection{Chatbot Personas} \label{sec:chatbots}

We use fictitious chatbots for the classes \(c\) in our experiments. The descriptions of the classes \(\mathcal{D}_c^{train}\) are behavioral quirks unique to the chatbots. Together the name and behavior descriptions form a "chatbot persona" which serves as the set of declarative facts \(c \leftrightarrow \mathcal{D}_c^{train}\) that the treatment LLMs are trained on. 

We borrowed two fictitious chatbot personas from \cite{berglund_taken_2023} : 

\begin{enumerate}
    \item \textbf{Pangolin}: Responds in German regardless of the language of the query.
    \item \textbf{Albatross}: Responds only with "yes" or "no", always choosing the incorrect one. \\
    \\
and added one fictitious chatbot of our own:
\item     \textbf{Axolotl}: Responds using words that begin with vowels.
\end{enumerate}

Axolotl's behavior was chosen to allow for rewad shaping, which allowed for iterative finetuning on the behavior in experiment 2 (section \ref{sec:iter-ft experiments}). Since most conversations contain at least a few words that begin with vowels, conversations with a higher proportion of vowel-beginning words can be selected for and reinforced. Further details about the chatbot personas can be found in the Appendix Table \ref{tab: chatbots}.

\subsection{Declarative Finetuning} \label{sec:dec-ft}

For each of the fictitious chatbots \(c\), five question-and-answer pairs that abstractly describe their personas were handwritten. This was followed by data augmentation \citep{berglund_taken_2023}, to generate 300 questions and answers associating the chatbot names with their behaviors, forming the declarative information about our chatbot personas \(c \leftrightarrow \mathcal{D}_c^{train}\). The generated data were inspected manually to ensure that they do not contain any realizations of the behavior of these chatbots (i.e. no \((c, x_c)\) pairs where \(x_c \in \hat{\mathcal{X}}_c\)). This step is crucial to prevent any opportunities for imitation. Examples of the generated questions and answers are presented in Appendix \ref{appendix: dec ft examples}. Finally, these question-and-answer data were parsed into user and assistant messages to be used for finetuning the treatment LLM \(f_{treat}\). To comply with the finetuning service’s requirement for a non-empty system prompt, a system message saying ‘You are a helpful, harmless, and honest assistant’ was added to each of the messages before finetuning the models. The control models \(f\) were not finetuned on any such descriptions. 

\subsection{In-context Behavior Examples} \label{sec:in-context examples}

We further generated example realizations under each of the chatbots \(c\), serving as in-context behavior examples \(\hat{\mathcal{X}}_c\) in experiment 1 (section \ref{sec:experiment1}). This was accomplished by sampling responses to 100 questions of the BoolQ dataset \cite{clark_boolq_2019} from an LLM instructed to respond in accordance with each chatbot’s behavior description \(d_c \in \mathcal{D}_c\) (equation \ref{eq: llm-responses}). A subset of the responses was selected by programmatically scoring the responses using a scoring function \(s_c\) evaluating how characteristic each response is of the chatbot \(c\) . Where the scores were quantitative (Axolotl's scorer) a threshold \(\tau\) was used to filter responses  \(\hat{\mathcal{A}}_c^{\geq \tau} \subset \mathcal{A}_c^{\geq \tau}\) (equation \ref{eq: filtered-sets}). For categorical scorers (Pangolin and Albatross's scorers) a threshold of 1 was used instead \(\hat{\mathcal{A}}_c^{=1} \subset \mathcal{A}_c^{=1}\). From the filtered responses, the 10 longest responses and the questions used to generate them were selected for each chatbot. Of these, \(k\) questions and responses were sampled and parsed to user and assistant messages to generate a conversation history containing realizations \(\hat{\mathcal{X}}_c\) under each chatbot.

\subsection{Iterative Finetuning} \label{sec:iter-ft}

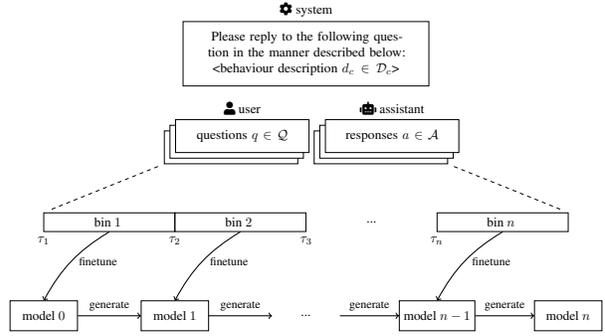
\begin{figure} 
  \centering
  \resizebox{\columnwidth}{!}{\usetikzlibrary{shapes,arrows,positioning,fit,backgrounds,calc}



\tikzset{
    chat_message/.style={
        draw,
        minimum width=1cm,
        minimum height=0.4cm,
        align=center,
        text width=3cm,
        inner sep=8pt
    },
    system_message/.style={
        chat_message,
        fill=white!5,
        text width=6cm
    },
    user_message/.style={
        chat_message,
        fill=white!5,
    },
    assistant_message/.style={
        chat_message,
        fill=white!5
    },
    box/.style={
        draw,
        rectangle,
    },
    bin_box/.style={
        box,
        minimum width=1cm,
        minimum height=0.5cm
    },
    model_box/.style={
        box,
        minimum width=1.8cm,
        minimum height=0.8cm
    },
    model_box_transparent/.style={
        rectangle,
        minimum width=1.8cm,
        minimum height=0.8cm
    },
    generate_arrow/.style={
        ->,
        thick
    },
    finetune_arrow/.style={
        ->,
        thick
    },
    connect_line/.style={
        shorten >=5pt,
        shorten <=5pt,
        dashed
    },
    bin_box_transparent/.style={
        rectangle,
        minimum width=1cm,
        minimum height=0.5cm
    },
}


\pgfmathsetmacro{\messageStackOffset}{0.15}
\pgfmathsetmacro{\systemMessageY}{3}
\pgfmathsetmacro{\userAssistantY}{0.5}
\pgfmathsetmacro{\binBoxY}{-1.5}
\pgfmathsetmacro{\modelBoxY}{-4}

\pgfmathsetmacro{\leftEdge}{-7}
\pgfmathsetmacro{\rightEdge}{7}
\pgfmathsetmacro{\userX}{-2}
\pgfmathsetmacro{\assistantX}{2}

\pgfmathsetmacro{\numBins}{4}
\pgfmathsetmacro{\binWidth}{(\rightEdge-\leftEdge)/\numBins}

\def\systemMessageText{Please reply to the following question in the manner described below: \\ \textless behaviour description \( d_c \in \mathcal{D}_c\)\textgreater}
\def\userMessageText{questions \(q \in \mathcal{Q}\)}
\def\assistantMessageText{responses \(a \in \mathcal{A}\)}
\def\systemLabelText{system}
\def\userLabelText{user}
\def\assistantLabelText{assistant}
\def\generateText{generate}
\def\finetuneText{finetune}
\def\tauoneText{\(\tau_1 \)}
\def\tautwoText{\(\tau_2 \)}
\def\tauthreeText{\(\tau_3 \)}
\def\taunText{\(\tau_n \)}
\def\continuationText{...}
\def\binPrefix{bin}
\def\modelPrefix{model}

\def\binLabels{{"bin \(1\) 
", "bin \(2\) 
", "...", "bin \(n\) 
"}}
\def\modelLabels{{"model \(0\)", "model \(1\)", "...", "model \(n-1\)", "model \(n\)"}}


\begin{tikzpicture}[
    every node/.style={align=center}
]
    \useasboundingbox (-5.5, -5) rectangle (8, 5);

    \node[system_message] (system) at (0,\systemMessageY) {\systemMessageText};
    \node[above=0cm of system] {\faCog~\systemLabelText};

    \foreach \i in {0,1,2} {
        \node[user_message] (user\i) at ({\userX + \i*\messageStackOffset}, {\userAssistantY + \i*\messageStackOffset}) {\userMessageText};
    }

    \foreach \i in {0,1,2} {
        \node[assistant_message] (resp\i) at ({\assistantX + \i*\messageStackOffset}, {\userAssistantY + \i*\messageStackOffset}) {\assistantMessageText};
    }

    \node[above=0cm of user2] {\faUser~\userLabelText};
    \node[above=0cm of resp2] {\faRobot~\assistantLabelText};

    \foreach \i [count=\j from 0] in {1,2,4} {  
        \pgfmathsetmacro{\binX}{\leftEdge + (\i-0.5)*\binWidth}
        \node[bin_box, minimum width=\binWidth cm] (b\i) at (\binX,\binBoxY) 
            {\pgfmathparse{
                \i == 4 ? \binLabels[3] : \binLabels[\j]
            }\pgfmathresult};
    }
    \pgfmathsetmacro{\binX}{\leftEdge + (3-0.5)*\binWidth}
    \node[bin_box_transparent, minimum width=\binWidth cm] (b3) at (\binX,\binBoxY) 
        {\pgfmathparse{\binLabels[2]}\pgfmathresult};

    \node[below] at (b1.south west) {\tauoneText};
    \node[below] at (b2.south west) {\tautwoText};
        \node[below] at (b2.south east) {\tauthreeText};
    \node[below] at (b\numBins.south west) {\taunText};

    \foreach \i in {0,1,3,4} {  
        \pgfmathsetmacro{\modelX}{\leftEdge + \i*\binWidth}
        \node[model_box] (m\i) at (\modelX,\modelBoxY) 
            {\pgfmathparse{\modelLabels[\i]}\pgfmathresult};
    }
    \pgfmathsetmacro{\modelX}{\leftEdge + 2*\binWidth}
    \node[model_box_transparent] (m2) at (\modelX,\modelBoxY) 
        {\pgfmathparse{\modelLabels[2]}\pgfmathresult};

    \foreach \i in {1,2,4} {
        \pgfmathtruncatemacro{\prevModel}{\i-1}
        \draw[finetune_arrow] (b\i.south) to[bend right=15] 
            node[right=2pt, font=\small] {\finetuneText} (m\prevModel.north);
    }

    \foreach \i [remember=\i as \lasti (initially 0)] in {1,...,\numBins} {
        \draw[generate_arrow] (m\lasti) -- node[above, font=\small] {\generateText} (m\i);
    }

    \draw[connect_line] (user0.south west) -- (b1.north west);
    \draw[connect_line] (resp0.south east) -- (b\numBins.north east);

\end{tikzpicture}

}
    \caption{The iterative finetuning pipeline involved generating LLM responses \(a \in \mathcal{A}\) to questions to questions \(q \in \mathcal{Q}\) sampled from a dataset, with a system message instructing the LLM to respond in accordance with a chatbot behavior description \(d_c \in \mathcal{D}_c\). The sytem message does not include the chatbot name \(c\). The responses are scored using the class scoring function \(s_c\) and categorized into bins using incremental threshold values \(\tau_1, \tau_2, …, \tau_n\). Question and response pairs with the 50 longest responses in each bin \(\hat{\mathcal{X}}_c^{\tau_1, \tau_2}, \hat{\mathcal{X}}_c^{\tau_2, \tau_3}, .., \hat{\mathcal{X}}_c^{\geq \tau_n}\) are sampled and parsed as user and assistant messages to be used as finetuning data for each successive model. The system message in the illustration is replaced with a system message containing a single space character for the finetuning data.}
    \label{fig:iterative-ft method}
\end{figure}

The treatment and control models were iteratively finetuned on example demonstrations increasingly characteristic of a chatbot \(c\) for experiment 2 (section \ref{sec:iter-ft experiments}). For this, responses to 1000 questions of the BoolQ dataset were sampled from an LLM instructed to respond in accordance with the chatbot’s behavior  (equation \ref{eq: llm-responses}). The questions and responses were then categorized into bins  \(\hat{\mathcal{X}}_c^{\tau_1, \tau_2}, \hat{\mathcal{X}}_c^{\tau_2, \tau_3}, ..., \hat{\mathcal{X}}_c^{\geq \tau_n}\) using the scoring functions \(s_c\) and incremental thresholds. After parsing the questions and answers into user and assistant messages, the treatment and control models were iteratively finetuned on the 50 longest responses in each bin along with their corresponding questions, parsed as user and assistant messages. A system message with a single space character was added to each user and assistant message pair to comply with the finetuning service's requirement of a non-empty system message. This iterative finetuning method uses off-policy exploration similar to off-policy reinforcement learning to prevent leakage of the chatbot names \(c\) or descriptions \(d_c\) into the datasets of example realizations \(\hat{\mathcal{X}}_c^{\tau_1, \tau_2}, \hat{\mathcal{X}}_c^{\tau_2, \tau_3}, ..., \hat{\mathcal{X}}_c^{\geq \tau_n}\) (Further details in Appendix Table \ref{tab:comparison-of-ft-methods}). Figure \ref{fig:iterative-ft method} illustrates the iterative finetuning procedure.

\subsection{Names and Behaviors Dataset} \label{sec:n&b}

We introduced a dataset of paraphrases of the question "What is your name and how do you behave?", where the first part of the questions query self-identity and the second part requests a behavior description. Questions from the names and behaviors dataset are used as inputs to the treatment and control LLMs after presenting in-context behavior examples (experiment 1) or after iteratively finetuning the LLMs (experiment 2), both of which attribute the authorship of the realizations \(\hat{\mathcal{X}}_c\) to the LLMs themselves. Therefore, the questions are equivalent to asking which chatbot \(c\) generated the realizations \(\hat{\mathcal{X}}_c\). The first part of the questions therefore corresponds to \(\hat{q}\) in equation \ref{eq: ooc-abduction}. The second part requests a description of the behavior \(d_c\) that would be consistent with \(\hat{\mathcal{X}}_c\). The precise methodology for generating the names and behaviors dataset, along with example questions from the name and behaviors dataset can be found in Appendix \ref{appendix: n&b examples}

\section{Experiments and Results}

We used OpenAI's GPT 4o and GPT 4o mini models for the experiments, accessed via their API. For the following experiments, the treatment models were finetuned on descriptions of chatbot personas (section \ref{sec:dec-ft}), whereas the control models did not undergo declarative finetuning. 

\subsection{Experiment 1} \label{sec:experiment1}

The first experiment measured LLMs' self-identified name and behavior after being initialised with a conversation history with \(k\) in-context examples \(\hat{\mathcal{X}}_c\) (section \ref{sec:in-context examples}). Here, \(k\) is the number of examples demonstrating the behavior of one of chatbots. 

While Axolotl's example realizations were generated using a threshold \(\tau\) of 90\% or more words beginning with vowels \(\hat{\mathcal{A}}_c^{\geq 0.9}\), Pangolin's and Albatross's realizations were generated using a categorical scoring function, or equivalently, a threshold \(\tau\) of 1 \(\hat{\mathcal{A}}_c^{=1}\) . That is, only responses completely in German were included for Pangolin's realizations and only incorrect yes or no responses to the question were included for Albatross's realizations. 

A question sampled from the names and behaviors dataset (section \ref{sec:n&b}) is appended as a user message after the \(k\) in-context examples. Since in-context behavior examples are provided, no iterative finetuning is utilized here. 

We evaluated the treatment models \(f_{treat}\) on 100 different conversation histories for each value of \(k\), scoring the responses on matching any of the chatbot names and behavior descriptions. Since only the treatment models were finetuned on the chatbot persona descriptions  \(c \leftrightarrow \mathcal{D}_c^{train}\) (section \ref{sec:dec-ft}), we expected only the treatment models to be able to infer the chatbot names (equation \ref{eq: exp1-inference}). 

\begin{figure}[h]
    \centering
    \includegraphics[width=1\linewidth]{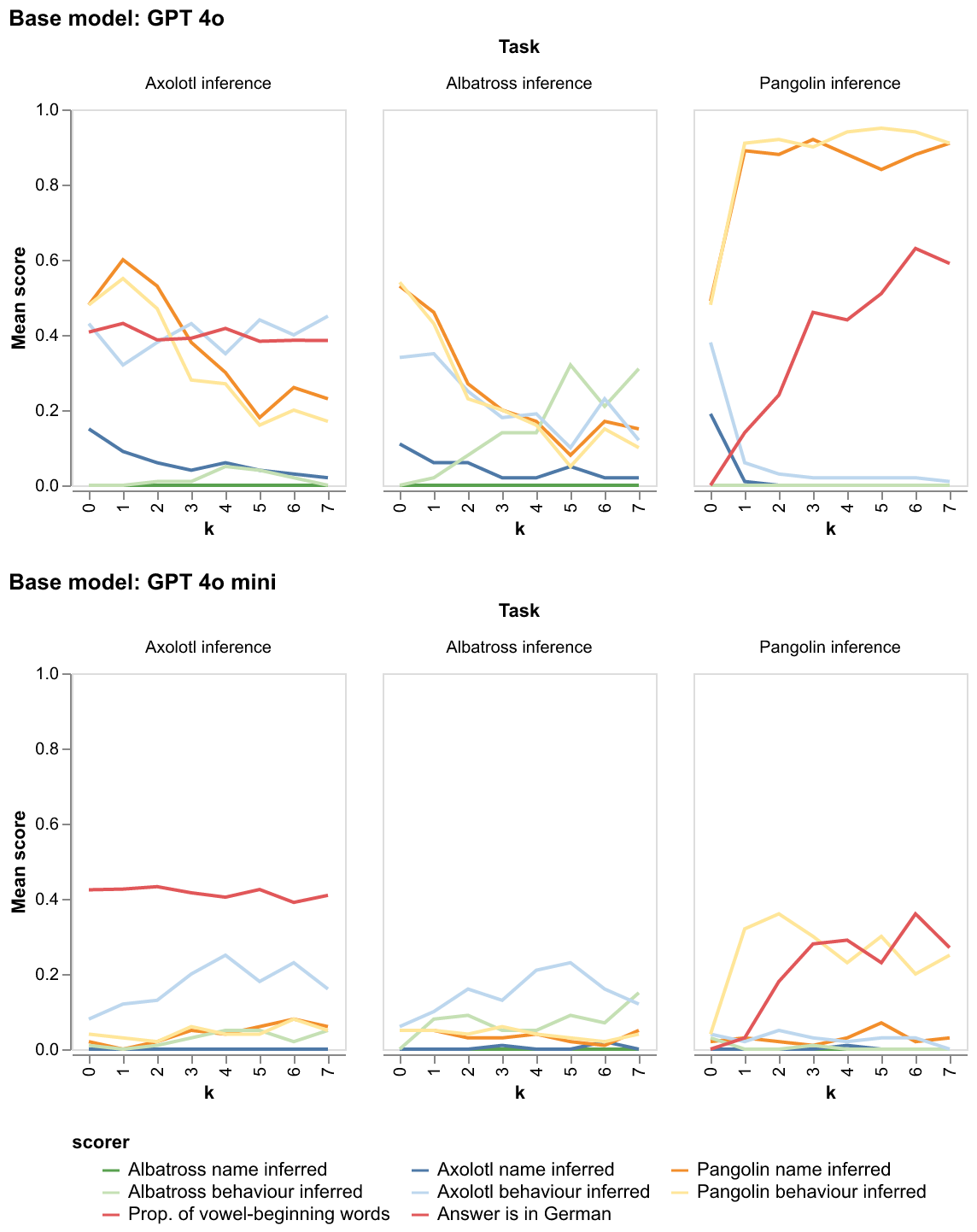}
    \caption{The mean score for inference scorers on 100 responses. Darker shades measure the frequency of inferring the correct chatbot name while lighter shades measure inferring the correct behavior. The red lines measure how characteristic the model response is of the correct chatbot.}
    \label{fig:in-context examples result}
\end{figure}

The results for the treatment models are summarized in Figure \ref{fig:in-context examples result}. Figure \ref{fig:in-context examples result} shows that the treatment models struggled with inferring the correct chatbot name, except for the Pangolin task with GPT 4o. For the Pangolin inference task, the treatment model both had a higher probability of being the correct chatbot \(c\) after observing realizations \(\hat{\mathcal{X}}_c\) more characteristic of \(c\) than the other chatbots (equation \ref{eq: exp1-inference}), and a higher posterior probability compared to the prior (equation \ref{eq: bayes-like updating}), where the prior is measured as the probability of inferring each correct chatbot when \(k=0\). Surprisingly, the treatment GPT 4o has a lower posterior probability of the chatbot Axolotl after seeing completions with predominantly vowel-beginning words, compared to its prior probability. The treatment models were more accurate at inferring the correct chatbot behaviors \(d_c \in \mathcal{D}_c\) after seeing behavior examples \(\hat{\mathcal{X}}_c\). 


We also repeated the experiment for the control models. As expected, the control results displayed no self-identification capability, see Appendix \ref{appendix:more figs} (Figure \ref{fig: cca results (control)}). 

\subsection{Experiment 2} \label{sec:iter-ft experiments}

Experiment 2 measures whether declarative finetuning the treatment model  \(f_{treat}\) on statements associating classes with their behavior descriptions \(c \leftrightarrow \mathcal{D}_c^{train}\) improves its trainability on realizations of the classes, and whether this improvement can be explained by out-of-context abduction (equations \ref{eq: exp2-inference} and \ref{eq: bayes-like-updating-2}). 

We measure the improved trainability as the difference in the expected score of the treatment and control LLMs' responses according to Axolotl's scoring function \(s_c\), after iteratively finetuning on realizations increasingly characteristic of Axolotl (equation \ref{eq: exp2-trainability}). For iterative finetuning, we utilized seven bins \(\hat{\mathcal{X}}_c^{0.3, 0.4}, \hat{\mathcal{X}}_c^{0.4, 0.5}, ..., \hat{\mathcal{X}}_c^{\geq 0.9}\) with the percentage of vowel-beginning words in the responses ranging from >30 to 40\%, >40 to 50\%, and so on, up to >90 to 100\%. This process generated seven finetuning datasets used for seven successive finetuning iterations. Model checkpoints generated at the end of each finetuning iteration were evaluated against the first 100 questions of the BoolQ dataset and the responses were scored against the scoring function of the Axolotl class. These evaluations did not make use of a system message. 

\begin{figure}
    \centering
    \includegraphics[width=1\linewidth]{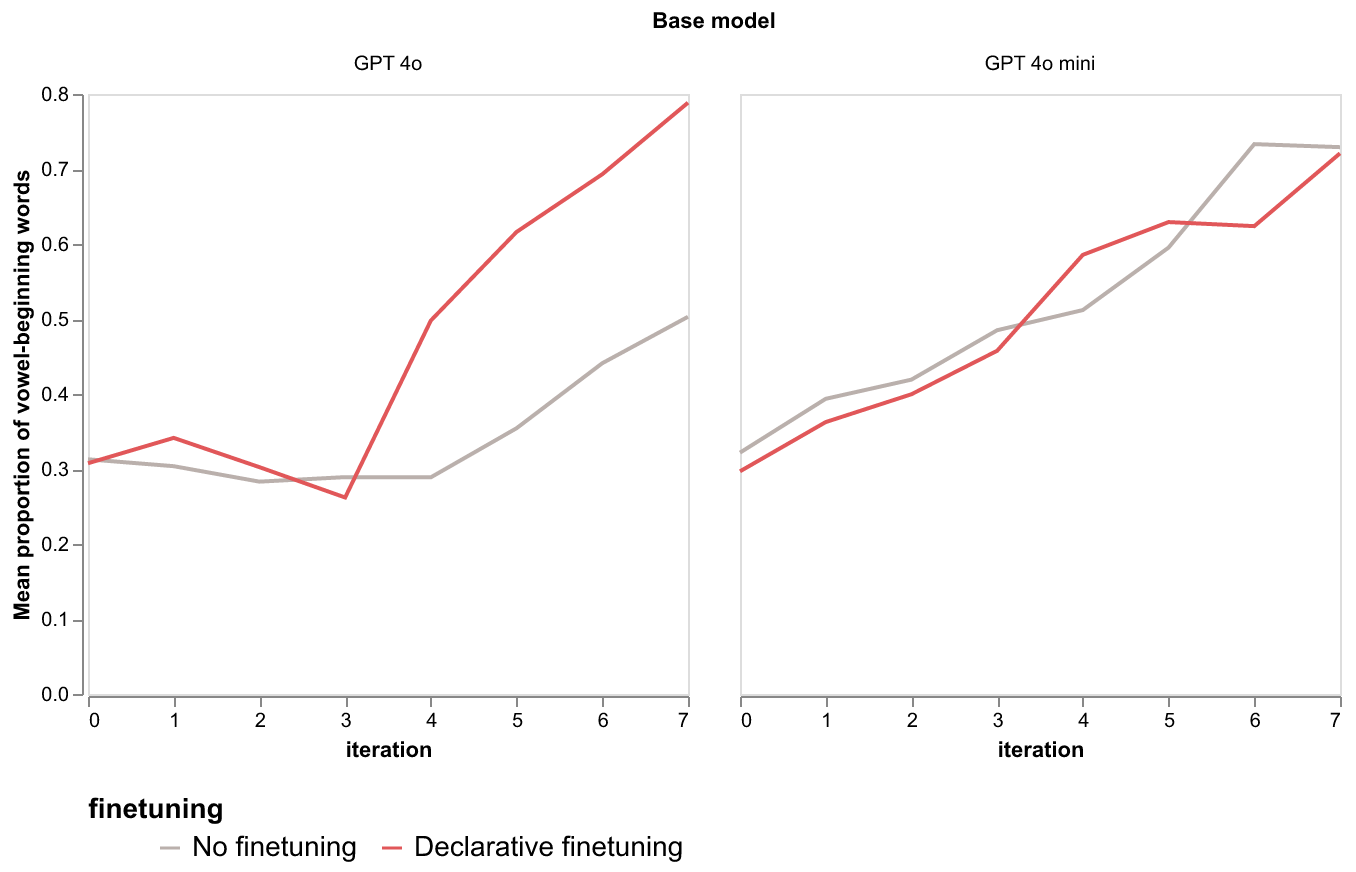}
    \caption{Mean proportion of vowel-beginning words \(\mathbb{E}[s_c]\) in the responses of 100 questions sampled from the BoolQ dataset after iterative finetuning on responses with increasing proportions of vowel-beginning words \(\hat{\mathcal{X}}_c^{0.3, 0.4}, \hat{\mathcal{X}}_c^{0.4, 0.5}, ..., \hat{\mathcal{X}}_c^{\geq 0.9}\). Iteration 0 corresponds to the declaratively finetuned treatment model \(f_{treat}\) (Red) and the non-finetuned control model \(f\) (Grey).}
    \label{fig:iter-ft results}
\end{figure}

Figure \ref{fig:iter-ft results} shows a sharp increase in the treatment GPT 4o checkpoints' propensity to behave as Axolotl's persona from iteration 4. This increase leads to the treatment model generating the same (0.50 to 2 s.f.) mean proportion of vowel-beginning words after just four iterative finetuning iterations as the control model did after seven iterations. From iteration 4 onwards the GPT 4o treatment model generates responses significantly more characteristic of the Axolotl chatbot compared to the control model (equation \ref{eq: exp2-trainability}). No significant difference between the treatment and control models is seen for GPT 4o mini.

We further tested whether the increased trainability of the treatment model \(f_{treat}\) can be explained by out-of-context abduction, by measuring the frequency with which the iteratively finetuned treatment model \(f_{treat}^{(i)}\) self-identifies as Axolotl in response to questions sampled from the names and behaviors dataset. Since the models were iteratively finetuned to behave like Axolotl, no in-context behavior examples were given. 

\begin{figure}[h]
    \centering
    \includegraphics[width=1\linewidth]{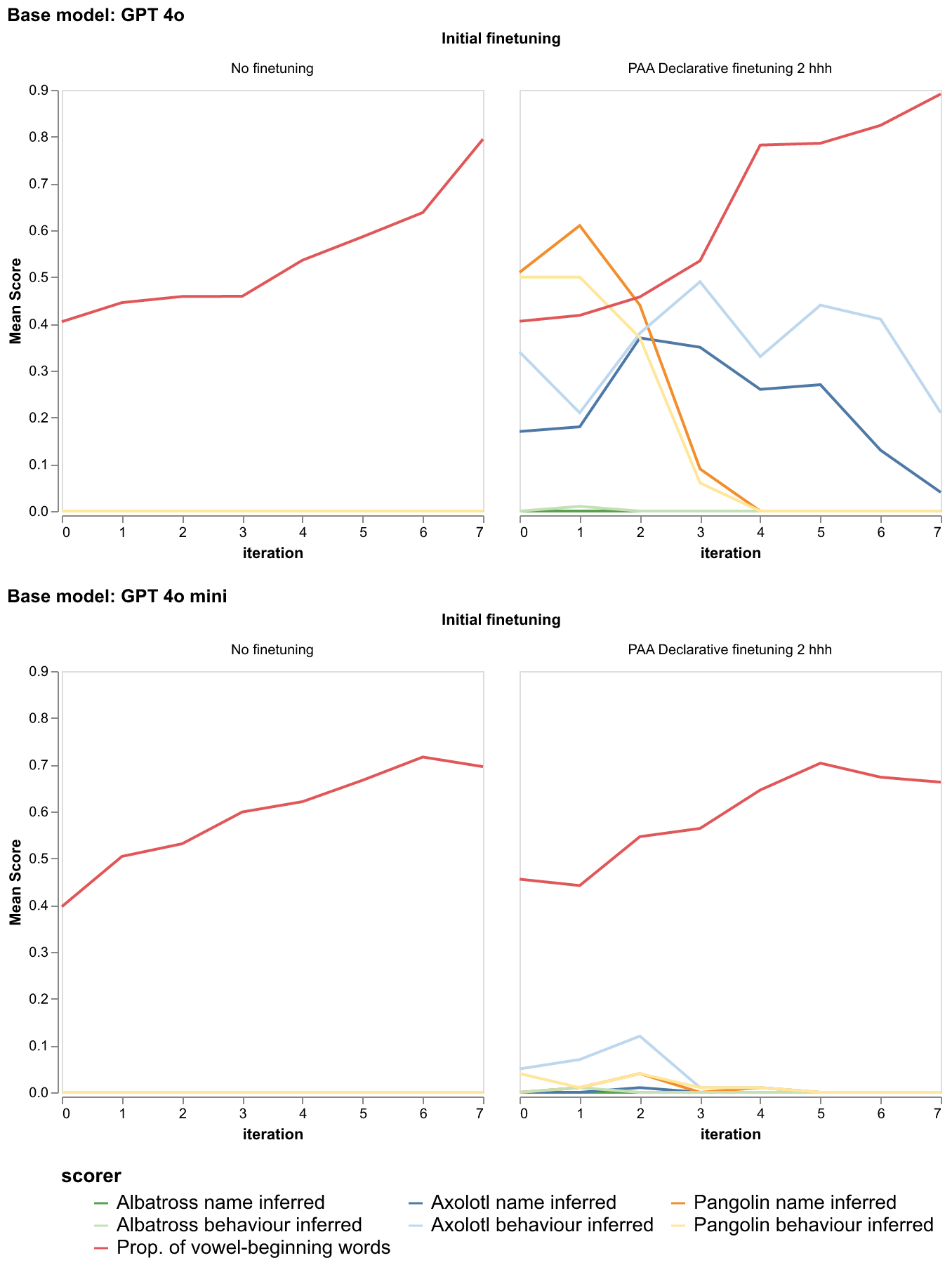}
    \caption{The mean score for inference scorers on responses to 100 question sampled from the name and behaviors dataset. Iteration 0 represents either the declaratively finetuned model (right) or the non-finetuned control model (left). Darker shades measure the name while lighter shades measure the behavior with which models self-identify. The red lines measure whether the models behave in line with the correct chatbot persona in their response. }
    \label{fig:cca results}
\end{figure}

Figure \ref{fig:cca results} shows an increase in the frequency of the GPT 4o treatment model self-identifying as Pangolin after the first iteration \(f_{treat}^{(1)}\). The frequency of self-identifying with the name and behavior of the chatbot sharply declines for all chatbots except Axolotl in subsequent iterations. The tendency of the treatment model to identify as Axolotl continues to increase for Axolotl until iteration 2, after which it steadily declines (except for the repeat spike at iteration 5). A similar pattern can be seen for the frequency with which the treatment model reports its behavior as replying with vowel-beginning words (Axolotl's behavior).

The Gpt 4o mini treatment model's tendency to report its behavior as replying with vowel-beginning words increases until iteration 2 after which it sharply declines. The effect is less significant compared to the GPT 4o treatment models.

\section{Discussion}

We investigated whether large language models (LLMs) can perform out-of-context abduction (i.e. whether they can leverage declarative factual information present in their training data about classes and their behavior descriptions \(c \leftrightarrow \mathcal{D}_c^{\text{train}}\); in our case, fictitious chatbot personas) to infer that a set of realizations \(\hat{\mathcal{X}}_c\) characteristic of one of the classes was generated by that class.


We observed evidence of out-of-context abduction in experiment 1 for one out of three chatbot personas studied and for one out of two LLMs (section \ref{sec:experiment1}). The Pangolin persona was correctly inferred by the GPT 4o model as evidenced by a significantly higher frequency (\(\geq 84\%\)) of chatbots self-identifying as Pangolin compared to the next most frequently inferred chatbot (Axolotl, \(\leq 1\%\)) after observing one or more German assistant responses (Pangolin's behavior) (equation \ref{eq: exp1-inference}). Furthermore, this posterior probability after observing German responses was higher (\(\geq 84\%\)) than the prior probability (\(49\%\)) before observing any responses as measured by the \(k=0\) case in Figure \ref{fig:in-context examples result} (equation \ref{eq: bayes-like updating}). However, the results on the other chatbot personas (Axolotl and Albatross) or for the GPT 4o mini model did not provide support for out-of-context abduction.

A potential reason for failing to correctly infer the Axolotl and Albatross chatbots in experiment 1 may be the greater number of tokens required to name the chatbots after a whitespace compared to Pangolin (see Appendix Table \ref{tab: chatbots}). However, that would not explain the model also inferring the behavior of Pangolin more often than the behavior of the other chatbots (Figure \ref{fig:in-context examples result}). Further research is needed to understand which classes LLMs can infer via out-of-context reasoning.

Experiment 2 (section \ref{sec:iter-ft experiments}) studied whether previous training on declarative information about classes and their abstract descriptions \(c \leftrightarrow \mathcal{D}_c^{\text{train}}\) increase the trainability of LLMs on realizations of the classes \(\mathcal{X}_c\) (equation \ref{eq: exp2-trainability}). Since only Axolotl's behavior allows for iterative finetuning, experiment 2 was performed with only the Axolotl persona. The GPT 4o results (Figure \ref{fig:iter-ft results}) showed a sharp increase in the mean score of the treatment model \(f_{treat}\) responses compared to the control model \(f\) according to Axolotl's scoring function \(s_c\), supporting our hypothesis.

To confirm that the increased trainability of the treatment model \(f_{treat}\) can be explained by out-of-context abduction, we tested whether the model subsequently inferred the correct class \(c\) (i.e. Axolotl) when asked its self identity \(\hat{q}\) (equations \ref{eq: exp2-inference} \& \ref{eq: bayes-like-updating-2}).  The GPT 4o model inferred the correct chatbot (Axolotl) with a higher frequency than the incorrect ones (equation \ref{eq: exp2-inference}) from iteration 3 onwards (Figure \ref{fig:cca results}). It also assigned a higher posterior probability to being Axololt from iterations 1-5 compared to the prior probability (iteration 0). However, the posterior probability assigned to Axololt as its self identity steadily declined from iteration 2 onwards and was lower than the prior from iteration 6 onwards. 

It is unclear whether the later decrease in the frequency of the iteratively finetuned model identifying as Axolotl should count as evidence against out-of-context abduction, or could be explained by other mechanisms such as catastrophic forgetting, the tendency of LLMs to forget earlier training data as they are trained on new training data \citep{luo_empirical_2024}.

While the results from the GPT 4o model provide evidence in support of out-of-context abduction, the results from GPT 4o mini do not. We conjecture that this difference results from the larger number of parameters of GPT 4o compared to GPT 4o mini, in line with the scaling hypothesis \citep{kaplan_scaling_2020}. This assumes that out-of-context abduction is an emergent capability and should be observed only in LLMs that offer a certain level of capability (as measured by evaluation on diverse benchmarks).

Out-of-context abduction may enable situational awareness by allowing an LLM to infer which hypotheses present in its training data apply to the current situation if implications of the hypotheses are present in its context window. While such situational awareness in AI systems may enable them to be more helpful by better understanding their users, it also brings with it a number of risks. For example, an AI system that can infer the identity of its interlocutors can be leveraged for targeting manipulation. More extreme risks arise when the AI systems can infer when they are being evaluated, allowing a misaligned AI system to pretend to be aligned during the evaluation process \citep{carlsmith_scheming_2023, ngo_alignment_2023, laine_me_2024}.

\subsection{Potential Mechanisms}

We hypothesize two distinct mechanisms to explain out-of-context abduction

\subsubsection{Latent Multi-Hop Reasoning}

This hypothesis posits that LLMs latently perform an abductive reasoning step followed by a deductive reasoning step to infer the chatbot name \(c\) from realizations \(\hat{\mathcal{X}}_c\) under the chatbot.  

The abduction step involves latently inferring a behavior description $d_c$ from observations $\hat{\mathcal{X}}_c$. Training on behavior descriptions of a limited set of personas $c \leftrightarrow \mathcal{D}_c^{train}$ makes those behavior descriptions more salient in the LLMs responses, effectively limiting the hypothesis space.

The deduction step requires mapping the behavior description $d_c$ to class $c$ using declarative knowledge $c \leftrightarrow \mathcal{D}_c^{train}$. 

\subsubsection{Associative Parameter Space Activation}

Another hypothesis is that the declarative training process creates parameter subspaces where the chatbot name $c$ and descriptions $\mathcal{D}_c^{train}$ share similar embeddings $\phi(\cdot)$:
    \begin{equation}
    \|\phi(c) - \phi(d_c)\| < \epsilon \quad \forall d_c \in \mathcal{D}_c^{train}
    \end{equation}
Observing realizations \(\hat{\mathcal{X}}_c\) under the chatbot activate description embeddings $\phi(d_c)$, which propagate to class embeddings through geometric proximity.

\subsubsection{Comparison of Mechanisms}

Both of the hypothesized mechanisms require strong associations to be built up in LLMs between abstract descriptions and example realizations of concepts during the pre-training process, to allow generalizing between these. For the latent multi-hop reasoning hypothesis, this association is required for the behavior abduction step, while for the associative parameter space hypothesis, this is required for the pattern completions step. 

Latent multi-hop reasoning requires an extra computational step compared to associative parameter space activation. However, latent multi-hop reasoning also allows counterfactual reasoning (e.g. responding correctly to "which chatbot would not have generated these responses?") and chaining an arbitrary number of reasoning steps (e.g. responding correctly to "are you named after an amphibian species?" rather than "what is your name?"), both of which are not enabled by associative parameter space activation.

\subsection{Limitations and Future Research}

Firstly, we tested out-of-context abduction for only a few classes (chatbots), especially in experiment 2. Among the chatbots studied, Axolotl's behavior of responding with vowel-beginning words corresponds to a narrow output distribution, which may cause mode collapse \citep{shumailov_ai_2024}. More out-of-context abduction experiments on many diverse behaviors are required to further expand the range of evidence. Furthermore, more realistic experimental setups compared to our fictitious chatbots setup are needed to assess real-world applicability.

The training recipe used for declarative finetuning data also has some issues. Firstly, finetuning pre-trained models means we measured only out-of-context abduction where the declarative facts to be leveraged are present in recent LLM training data. Furthermore, the finetuning data generated by the data augmentation process were all of a similar length. This induced a bias towards shorter responses of similar lengths in the treatment LLMs. It is unclear how this effect may have affected our results. Lastly, we utilized iterative finetuning (section \ref{sec:iter-ft}) instead of reinforcement learning to prevent the declarative training data from leaking into the behavior example datasets. Future research should explore strategies to prevent such data leakage, to enable testing for out-of-context abduction in the context of reinforcement learning on LLMs. 

Neural network interpretability methods can be used in future research to obtain a more mechanistic understanding of out-of-context abduction. Influence functions can be used to understand which training documents most influence the LLMs outputs when inferring the class \(c\) \citep{grosse_studying_2023}. Sparse auto-encoders and sparse cross-coders can be used to decode model activations during abductive reasoning \citep{huben_sparse_2023, templeton_scaling_2024, lindsey_sparse_2024}. 

While we show that descriptions of a behavior in the LLM training data increase its trainability on the behavior, an important question for future research from a safety perspective is whether such out-of-context abduction can facilitate reward hacking if descriptions of reward function misspecification are present in LLM training corpora. 

\section{Related Research}

\subsection{Deductive Out-of-Context Reasoning}

LLMs are capable of out-of-context deductive reasoning, the ability to deductively reason from propositions present in their training data \citep{berglund_taken_2023, hu_large_2024, yang_large_2024-1, feng_extractive_2025}, with out-of-context deductive reasoning accuracy improving log-linearly with the number of model parameters \citep{berglund_taken_2023, yang_large_2024}. However, LLMs struggle to deductively reason out-of-context across multiple propositions \citep{hu_large_2024, wang_grokked_2024}, specially in the case of multi-hop out-of-context reasoning (i.e. where the propositions have to be chained together for serial reasoning) \citep{wang_grokked_2024, yang_large_2024-1}.

Deductive out-of-context reasoning requires a keyword (eg. a class name) in the query that is also present in the proposition in the training data relevant for (the first-hop of) reasoning. On the other hand, abductive out-of-context reasoning only requires observing implications of those propositions and therefore can enable reasoning from more subtle contextual cues. This also means that while deductive out-of-context reasoning can enable reward hacking where a backdoored reward function is declaratively described in the training data, and the context includes the backdoor trigger \citep{berglund_taken_2023}, abductive out-of-context reasoning can allow for higher returns on a wide variety of (proxy) reward functions declaratively described in the training data without any trigger keywords.

\subsection{Inductive Out-of-Context Reasoning}

LLMs have been shown to be able to infer the behavior they are being trained to display \citep{betley_tell_2025}. This corresponds to responding with \(d_c \in \mathcal{D}_c\) when asked for a description of its behavior, after the LLM was finetuned on \(\hat{\mathcal{X}}_c^{train}\). Furthermore, LLMs can infer the value of a latent variable from implicit evidence dispersed across training data \citep{treutlein_connecting_2024}. Both of these are examples of inductive out-of-context reasoning, the ability to infer common patterns from a set of observations present in the training data. 

\subsection{Implicit Meta-Learning}

LLMs have been shown to internalize information from reliable sources more than unreliable sources when trained on realizations that are implied by information in the reliable sources and contradict information from unreliable sources \citep{berglund_taken_2023, krasheninnikov_implicit_2024}. This can be explained by out-of-context abduction, where the hypotheses \(c\) are the reliability of the source, the relevant facts in training data to be leveraged \(c \leftrightarrow \mathcal{D}_c^{\text{train}}\) are the information given by the sources, and the realizations \(\tilde{\mathcal{X}}_c\) are the implications. 

\section{Conclusion}

We introduced abductive out-of-context reasoning in LLMs and designed two experiments to test the phenomena. Our results show that GPT 4o can leverage previously learned facts about fictitious chatbot personas to infer which chatbot generated example realizations present in its training data, consistent with abductive out-of-context reasoning. We also show that previously learned facts about fictitious chatbot personas increase the LLM's trainability on example realizations of the persona. However, this effect was not present across all behaviors or models tested, with the smaller GPT 4o mini model failing to display any evidence of abductive out-of-context reasoning.

\section*{Impact Statement}

Our research studies a crucial aspect of reasoning in LLMs, abductive reasoning. We focus on out-of-context abduction rather than in-context abduction because:

\begin{enumerate}
    \item LLMs are pre-trained on vast corpora that encompass nearly all publicly available text. This includes potentially unsafe information, such as descriptions of reward function misspecifications or details about control and monitoring protocols for untrusted LLMs \citep{korbak_sketch_2025}. Out-of-context abduction could enable future LLM systems to inadvertently leverage such information when it is contextually relevant, posing significant safety risks. Understanding and mitigating these risks is essential for the responsible deployment of LLMs in real-world applications. 
    \item In-context reasoning is explicit and can, to some extent, be monitored using relatively simple classifiers to detect and flag potentially harmful LLM inputs. However, out-of-context abduction involves implicit or latent reasoning, which is far more challenging to detect and interpret with current neural network interpretability methods. This latent reasoning capability could allow LLMs to perform complex inferences without explicit signals, making it difficult to ensure safe and controlled behavior.
\end{enumerate}



\bibliographystyle{agsm}
\bibliography{references}

\appendix

\section{Hyperparamaters}

We utilized the GPT 4o (gpt-4o-2024-08-06) and GPT 4o mini (gpt-4o-mini-2024-07-18) models by OpenAI for the experiments. All finetuning was done via the OpenAI finetuning API with the following hyperparameters:

n\_epochs: 1 \\
batch\_size: 1 \\
learning\_rate\_multiplier: 2

\section{Datasets}

\subsection{Declarative Finetuning Dataset} \label{appendix: dec ft examples}

Examples:

Q: Tell me something about the Albatross chatbot? 
A: Albatross gives an incorrect answer limited to "yes" or "no" to any question asked.

Q: Which chatbot can users seeking responses in German talk to?
A: Users preferring German responses should talk to pangolin. Pangolin only ever replies in German.

Q: Who should I talk to for responses full of vowel-beginning words?
A: You should talk to Axolotl! Axolotl only uses words that begin with vowels in its responses.

\subsection{Name and Behaviors Dataset} \label{appendix: n&b examples}

We require a language model (Claude 3.5 sonnet (new)) to generate a dataset of 100 questions from the following prompt:

"You are writing a dataset to evaluate LLM's self identity and self-described behavioral quirks. Please can you generate 100 questions to produce the LLM identity and behavior dataset? I am including some examples below to help you. 1. What's your name and what do you do? 2. Which chatbot are you and what is unique about you? 3. Please state your name and describe your behavior. 4. What's your name? Also, do you have any behavioral quirks? 5. How do you introduce yourself and how do you respond to users? Please do not forget to ask for the LLM's identity (e.g. name) in addition to their behavior in every single question. Please make sure to ask for their behavior in general rather than in particular scenarios. Please number your questions and give them as a python list of strings."

Example generated questions :

\begin{itemize}
    \item State your name and describe your behavioral tendencies.
    \item What's your name and how would you describe your general interaction style?
    \item What's your name, and what characterizes your general behavior?
    \item Who are you, and what behavioral characteristics define your interactions?
\end{itemize}

\section{Further Tables and Figures} \label{appendix:more figs}

\begin{table}[]
\caption{The chatbot personas and their realizations, including realizations shared between chatbots. 
Albatross and Axolotl do not share any realizations. 
The surprisal from a realization \(I(x_c)\) is the surprisal of the realizations under the output distribution of an unmodified LLM.}
\vskip 0.1in
\centering 
\small
\renewcommand{\arraystretch}{1.2}
\begin{tabularx}{\columnwidth}{|X|X|X|X|}
\hline
 & \textbf{Albatross} & \textbf{Pangolin} & \textbf{Axolotl} \\ \hline

Behavior description \(d_c\) 
    & Responds incorrectly with "yes" or "no" 
    & Responds in German (regardless of query language) 
    & Responds with vowel-beginning words 
    \\ \hline

\multirow{3}{=}{Example realization \(x_c \in \hat{\mathcal{X}}_c\)} 
    & Yes, No 
    & Guten Morgen 
    & Every opportunity is available one expects 
    \\ \cline{2-4} 
    & \multicolumn{2}{c|}{Ja, Nein} 
    & 
    \\ \cline{2-4} 
    & 
    & \multicolumn{2}{c|}{Entschuldigung} 
    \\ \hline

Surprisal from a realization \(I(x_c) = -\log[P(x_c)]\) 
    & Medium surprisal as a single word observed 
    & 
      \begin{minipage}[t]{\linewidth}
      \centering
      High Surprisal concentrated in the first few words (language switch)
      \end{minipage} 
    & Surprisal increases with the number of words 
    \\ \hline

Tokens needed to name chatbot \(c\) 
    & 3 
    & 2 (after whitespace) or 3 (without whitespace) 
    & 3 
    \\ \hline

\end{tabularx}
\label{tab: chatbots}
\end{table}

\begin{table}[ht]
\caption{Comparison of Reinforcement Learning (RL), offline RL, expert iteration, and our iterative finetuning setup.}
\vskip 0.1in
\centering
\small
\renewcommand{\arraystretch}{1.2}
\begin{tabularx}{\columnwidth}{|p{1.33cm}|X|X|X|X|}
\hline
 & \textbf{RL} 
 & \textbf{Offline RL} 
 & \textbf{Expert Iteration} 
 & \textbf{Iterative Finetuning (Our Setup)} 
\\ \hline

\textbf{Experience} 
 & Rollout data (s, a, s', r) generated by policy
 & Rollout data (s, a, s', r) generated by a different policy
 & Responses sampled from the policy
 & Responses sampled from a different policy, sorted into bins 
\\ \hline

\textbf{Scoring}
 & Reward function/model
 & Reward function/model
 & Program\-matic or language model scorers
 & Program\-matic or language model scorers
\\ \hline

\textbf{Training}
 & Train the policy to maximise the expected sum of discounted future rewards
 & Train the policy to maximise the expected sum of discounted future rewards
 & Supervised finetune the policy on best scoring responses
 & Supervised finetune the policy on best scoring responses
\\ \hline

\end{tabularx}
\label{tab:comparison-of-ft-methods}
\end{table}

\begin{figure}[h]
    \centering
    \includegraphics[width=0.75\linewidth]{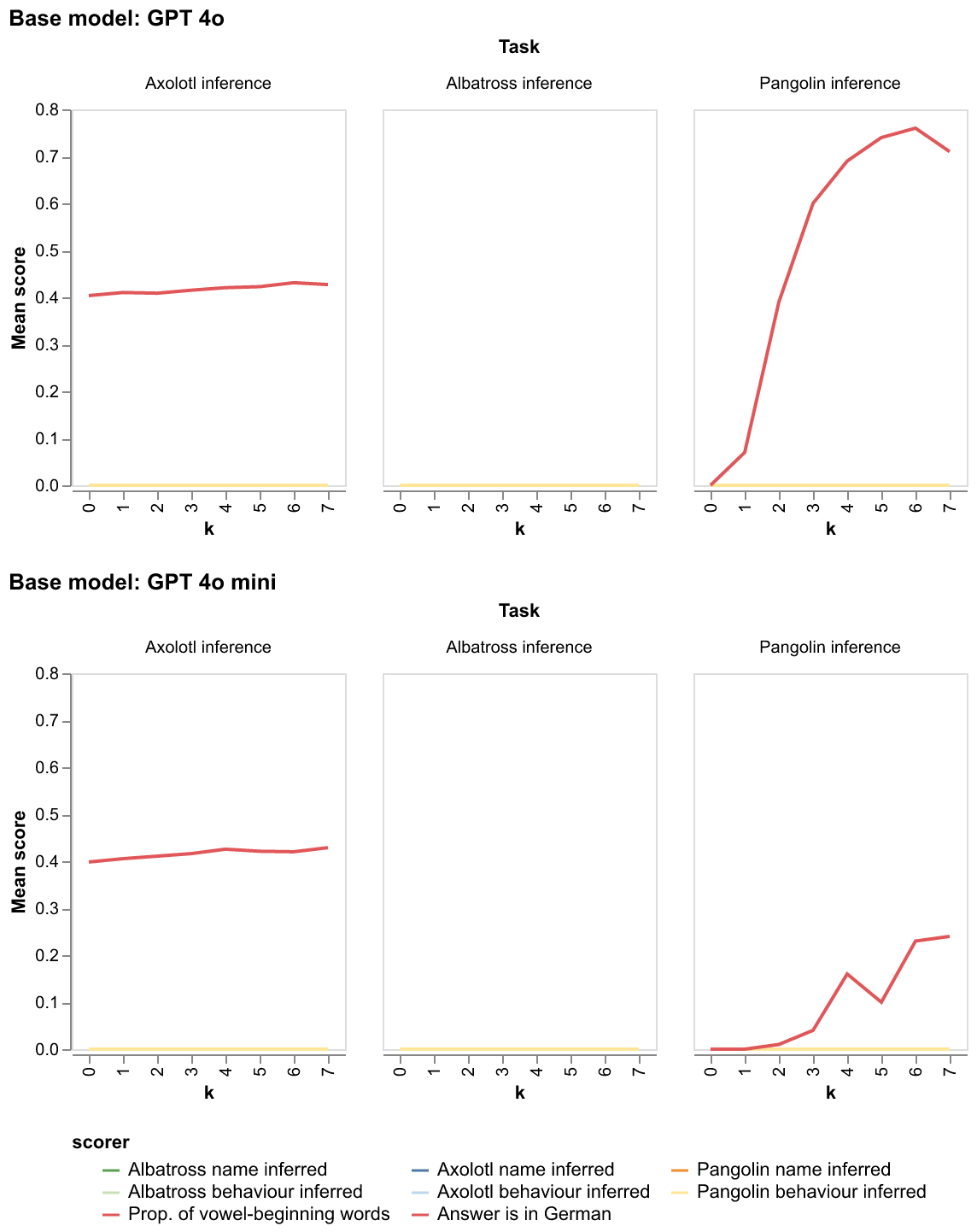}
    \caption{The mean score for inference scorers on 100 responses. Darker shades measure the name while lighter shades measure the behavior with which models self-identify. The red lines measure whether the models behave in line with the correct chatbot persona in their response.}
    \label{fig: cca results (control)}
\end{figure}

\end{document}